\documentclass[11pt,a4paper]{article}
\usepackage[hyperref]{emnlp2020}
\usepackage{times}
\usepackage{latexsym}
\usepackage{tikz}
\usepackage{booktabs} 
\usepackage{tabularx}
\def\checkmark{\tikz\fill[scale=0.4](0,.35) -- (.25,0) -- (1,.7) -- (.25,.15) -- cycle;}

\usepackage{microtype}

\aclfinalcopy 

\title{SLM: Learning a Discourse Language Representation with Sentence Unshuffling}

\author{Haejun Lee$^\heartsuit$ \qquad Drew A. Hudson$^\clubsuit$ \qquad Kangwook Lee$^\heartsuit$ \qquad  Christopher D. Manning$^\clubsuit$\\
  $\heartsuit$ Samsung Research \qquad  $\clubsuit$ Stanford University \\
  \texttt{\{haejun82.lee, kw.brian.lee\}@samsung.com} \\
  \texttt{\{dorarad, manning\}@cs.stanford.edu} \\}

\date{}

\begin{document}
\maketitle

\begin{abstract}
We introduce \textbf{S}entence-level \textbf{L}anguage \textbf{M}odeling, a new pre-training objective for learning a discourse language representation in a fully self-supervised manner. Recent pre-training methods in NLP focus on learning either bottom or top-level language representations: contextualized word representations derived from language model objectives at one extreme and a whole sequence representation learned by order classification of two given textual segments at the other. However, these models are not directly encouraged to capture representations of intermediate-size structures that exist in natural languages such as sentences and the relationships among them. To that end, we propose a new approach to encourage learning of a contextualized sentence-level representation by shuffling the sequence of input sentences and training a hierarchical transformer model to reconstruct the original ordering. Through experiments on downstream tasks such as GLUE, SQuAD, and DiscoEval, we show that this feature of our model improves the performance of the original BERT by large margins. 
\end{abstract}

\section{Introduction}
Recent representation learning methods in NLP such as BERT \cite{devlin2018bert} have focused on learning two types of representations: the bottom-level -- a contextual representation centered at a single word, trained by recovering randomly masked tokens or predicting previous and next words, and the top-level text, implicitly represented as a single [CLS] symbol and trained by predicting a relation between input segments, which usually consist of multiple sentences. However, natural language text has, in contrast, a very dominant hierarchical structure, with words grouped together into intermediate semantic units such as phrases and then sentences to convey the full meaning of a given text. Neither the transformer architecture nor the pre-training task leverages this hierarchy, treating the language as a flat sequence of tokens instead.

\begin{figure}[t]
\centering
\includegraphics[width=7cm]{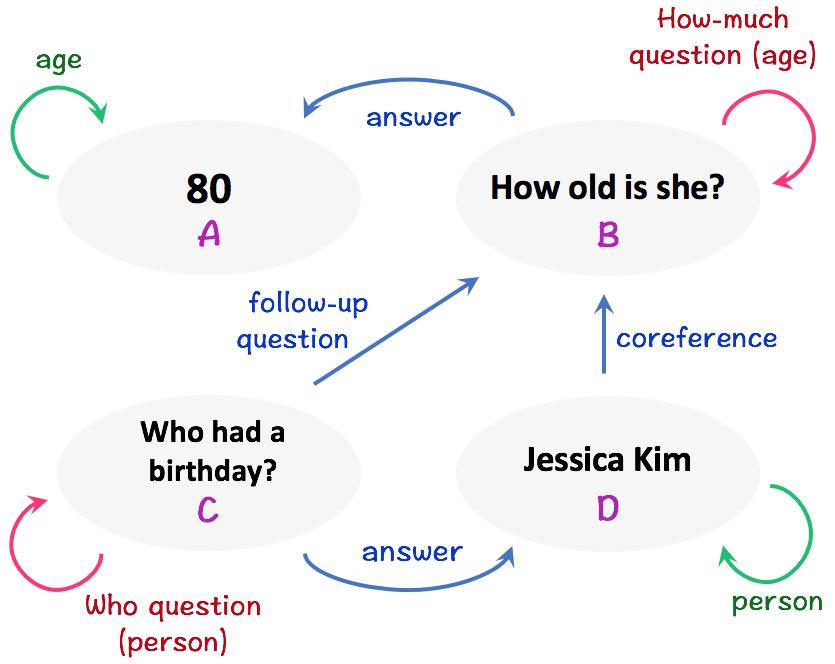}
\vspace{-0.1cm}
\caption{\label{conv} An example of a shuffled conversation and potential features that can aid in reconstructing the original sentence ordering, C $\rightarrow$ D $\rightarrow$ B $\rightarrow$ A.}

\vspace{-0.6cm}
\end{figure}

Inspired by prior works about sentence-based representations for recurrent networks \cite{kiros2015skip,hill2016learning,gan2017learning,jernite2017discoursebased, logeswaran2018sentence,gong2016end,chen2016neural}, we seek to incorporate a more explicit hierarchy into the transformer by extending it with the capacity to learn contextualized sentence-level representations. Equipping the model with such a capacity allows it to learn languages at multiple levels of granularity, ranging from fine-grain connections across words to high-level discourse relations between sentences and paragraphs.  

Predicting an original sequence of sentences requires deep understanding of natural language, including a variety of phenomena such as discourse relations, coreference, temporal dependencies, entailments, narratives, and so on. Figure~\ref{conv} shows a set of statements from a conversation in the CoQA dataset \cite{DBLP:journals/corr/abs-1808-07042}, and clues that can potentially be used to reconstruct their original ordering. To figure out the correct order, in this case C-D-B-A, a model has to understand that D and A are answers to questions C and B respectively and that `she' in B refers to D, and it may need to know that B is a follow-up question on C. Not only are the semantics of separate sentences important but the relationships between them are crucial in solving this task. We therefore seek to encourage the model to capture these vital properties by training it to reconstruct the original sentence ordering.

To achieve this, we propose a pre-training objective that extends the word-level language modeling to analogously learn representations for sentences. Typical word-level language models are trained by guessing the neighbors of given words or by predicting masked words based on their context. Applying the same idea at a higher level, our approach learns representations that support predicting the next sentence representation among shuffled sentences for given previous representations, based on the semantic relations between them. 

To allow the model to be effectively trained with the new unshuffling objective, we propose a pointer-based neural module that is specialized to predict the sentence order, called the Sequence Reconstructor (SR). In the SR, each sentence is represented by a sentential token that is inserted in front of it. A pointer network layer stacked on the transformer decoder is trained to point at the next contextualized sentence representation based on the previous sentence representations. 

We show that our method achieves robust improvements over standard BERT's performance on the following downstream NLP tasks: GLUE for Natural Language Inference \cite{DBLP:journals/corr/abs-1804-07461}, SQuAD for Question Answering \cite{rajpurkar2018know}, and DiscoEval \cite{mchen-discoeval-19} for discourse aware sentence representations. We match the score of Text-To-Text Transfer Transformer Base (T5\textsubscript{BASE}) \cite{Raffel2019ExploringTL}, a state-of-the-art model that uses BERT\textsubscript{BASE} hyperparameters, while using only half the parameters, shorter training (3/8 tokens overall), and a fraction (1/37) of the data compared to T5.
Moreover, we investigate the effect of the proposed objective through a qualitative analysis of the neighbor sentences of sentences that have similar sentential representations. We show that the results support our aim of enriching the transformer model with sentence-level language understanding. 

The contributions of our work are threefold: 1) we propose a new self-supervised pre-training objective that extends a word-level language modeling strategy to the sentence-level; 2) we propose a pointer-based neural module to train the model for this objective, and demonstrate it leads to significant improvements over diverse NLP tasks; and 3) we provide a qualitative analysis showing that the learned contextualized sentence representations embed more subtle and rich structural, semantic, and relational information.

\begin{figure}[t]
\includegraphics[width=7.5cm]{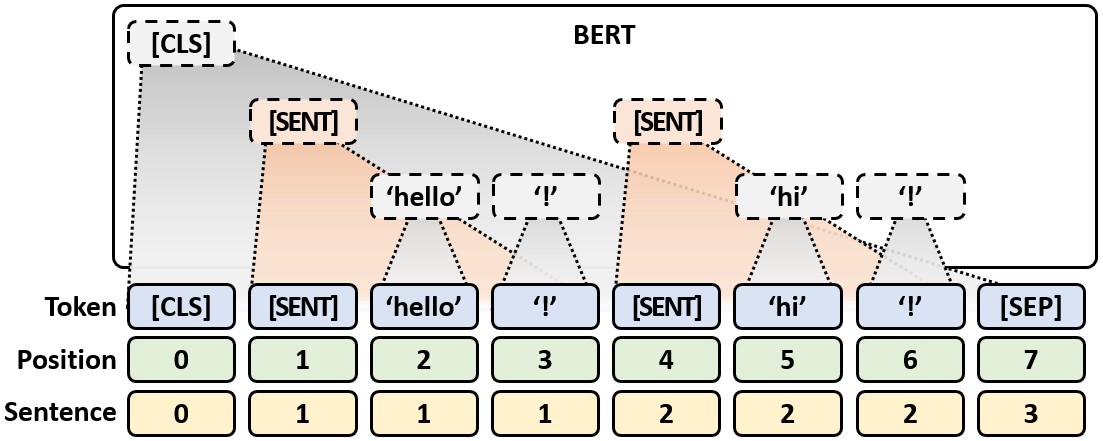} 
\caption{Intended representation scope of each contextualized sentence representation}
\label{representation_range}
\end{figure}

\begin{figure*}[t]
\includegraphics[width=16cm]{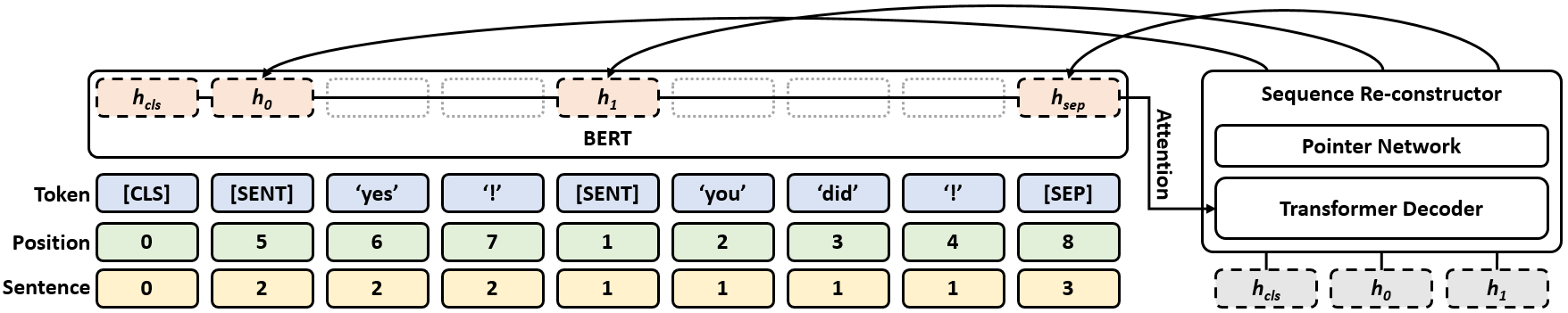}
\caption{\label{fig:architecture}  An overview of the architecture for the sentence unshuffling. Input sentences are randomly shuffled and the original ordering is reconstructed by our Sequence Reconstructor model using the sentence representations.}

\end{figure*}

\section{Proposed Method}
\label{method}
In this section, we describe the details of our proposed methods. A conventional representation model is extended with sentence representations to gather sentence information and a Sequence Reconstructor (SR) to predict their original ordering. Fine-tuning architectures are modified to inject sentence representation into downstream NLP tasks.

\subsection{Pre-training with Sentence Unshuffling}
\label{model_architecture}

Our model consists of an encoder for contextualizing input texts that uses a conventional transformer model like BERT, and a decoder that reconstructs the original ordering. We begin by splitting the input text into target segment units, sentences in this case, and shuffle them to remove their original ordering. We then add special tokens at the beginning of each segment, which will be used as sentence representations aggregating their meaning from their surroundings. We define a sentence-level language model (SLM) loss for reconstructing the original sentence ordering by making a sequence of predictions with a pointer network. Finally, the model is trained by sum of this loss and the standard masked language modeling (MLM) loss to pre-train the model. Through this pre-training objective, representations of sentences are properly contextualized by their neighboring sentence representations. Although we only consider sentence representations in this paper, our method can easily be extended to other syntactic levels such as phrases or paragraphs. We provide further detail about each of these components in the followings.

\vspace{1mm}

\noindent
\textbf{Sentence Representation:} We insert a sentence token, [SENT] in front of every sentence, that is meant to represent each sentence and aggregate its meaning from its surroundings. Similarly to positional and segment-level embeddings in the original BERT model, and to distinguish sentences and limit the scope of the sentence representations, we add trainable sentence embeddings indicating the index of each sentence to each of the word representations in order to support potential sentence-level aggregation as in ERNIE ~\cite{sun2019ernie}. Figure~\ref{representation_range} shows trainable embeddings added to inputs and the scope of input tokens that are represented by each contextualized representation. 

\vspace{1mm}

\noindent
\textbf{Sequence Shuffling:} Input sentences are shuffled by manipulating their positional and sentence embeddings (as the positional encoding is the only clue to derive the actual sequences in a transformer architecture, unlike traditional RNNs). For example, the inputs of Figure~\ref{fig:architecture} are a pair of swapped sentences. Position embeddings of the first sentence are 5 to 7 and those of the second sentence are 1 to 4. The sentence embedding are similarly swapped as well. Note that importantly the model can perform the unshuffling task only based on the semantics of the sentences and the relations between them, rather than by using the positional embeddings, since these are shuffled together with the sentences (namely, if sentence X and Y are swapped, so are their positional embeddings). For each iteration, only half of the batches are shuffled to allow the model to see some of the input in its natural ordering. 
\vspace{1mm}

\noindent
\textbf{Sequence Reconstructor:} SR is our conditional decoder consists of a pointer network and transformer decoder similar to \citet{gong2016end} and \citet{logeswaran2018sentence}. It predicts the original ordering of the shuffled input as depicted in Figure~\ref{fig:architecture}. After encoding the shuffled inputs by BERT, we consider the contextualized embeddings: $C=[h_{cls}, h_0, ..., h_{N-1}, h_{sep}]$ that correspond to the $N$ sentence tokens we have inserted into the input sequence, along with the first ([CLS]) and last ([SEP]) tokens. It is passed to the Sequence Reconstructor for the task of reconstructing the original sentence ordering. 

\begin{figure*}[t]
\includegraphics[width=16cm]{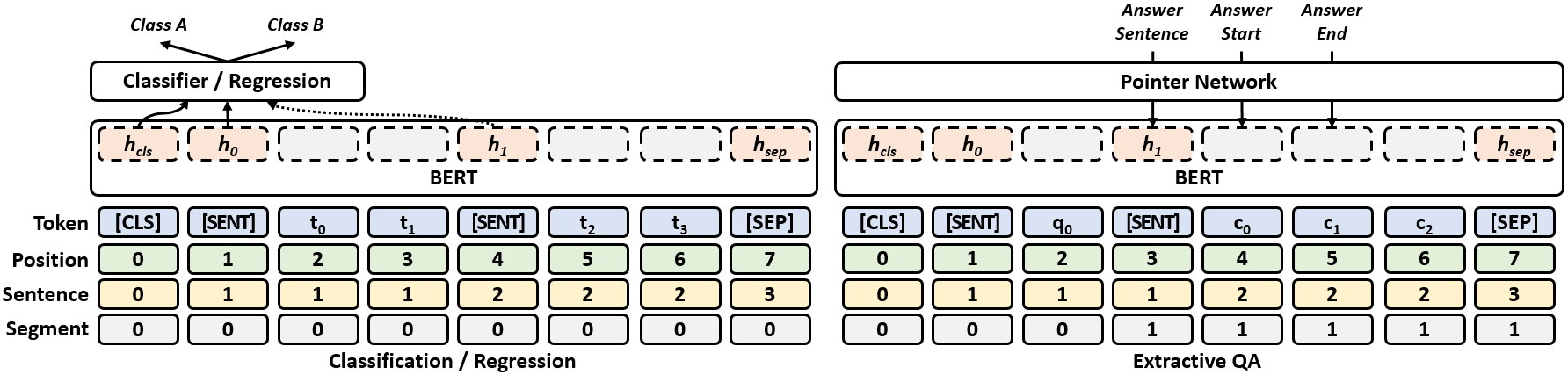} 
\caption{\label{finetune_arch} Fine-tuning Architectures. Sentence representations are concatenated to the [CLS] representation before being fed into the output layer for downstream classification (or regression) tasks and used to predict an answer sentence for extractive QA tasks. }

\end{figure*}

The embeddings in $C$, except the last $h_{sep}$, are sequentially processed through a transformer decoder \cite{transformer} to obtain output $\textit{w}_i$ while attending on the whole $C$ and embeddings processed in the previous steps, $\{C_j\}^{i-1}_{j=0}$. Each processed output will be used to predict the sentence $i$ in step $i$ (Note that indexes of sentence representations in $C$ are shifted by 1 due to $h_{cls}$ inserted in front). 

\begin{equation}
\label{formula_transforer_dec}
\mathrm{\textit{w}}_i=\mathrm{TransformerDecoder}(C_i,\{C_j\}^{i-1}_{j=0}, C) 
\end{equation}

Then, we compute a probability distribution over the sentence representations using a pointer network. Each probability represents the likelihood of each sentence to come up next at step $i$, aiming to predict their original ordering:

\begin{equation}
\label{formula_softmax}
\mathrm{P}_i=\mathrm{Softmax}(\mathrm{\textit{w}}_iC^T)
\end{equation}
We calculate a cross-entropy loss to match for each contextualized sentence embedding the probability distribution of the predicting ordering with the golden positions. $\mathrm{P}_i = {\{p_{i,j}\}}_{j=0}^{N+1}$ is the predicted probabilities of sentence $j$-1 (due to a shift by $h_{cls}$) appearing at position $i$, and $\mathrm{o}_{i}$ is a one hot ground-truth vector of the correct position at step $i$. Finally, we average the losses for all positions $i=0,\dots,N$: 
\begin{equation}
\label{formula_loss}
\mathrm{\mathcal{L}_{slm}}=-\frac{1}{N+1}\sum^{N}_{i=0}\displaystyle\sum^{N+1}_{j=1} \mathrm{o}_{i,j} \log{(p_{i,j})}
\end{equation}
Intuitively, the decoder learns how to transform a representation of sentence to the next sentence's representation recurrently by attending on the whole sentence representations from the encoder and previously transformed embeddings.

One important design choice here is that the Sequence Reconstructor only sees $C$, which are contextualized embeddings of [CLS], the sentence tokens, and [SEP] while other contextualized word representations are masked out. The model should find out the original sequence using only sentence representations and this will enforce the encoder to embed all necessary information into the contextualized sentence embedding instead of spreading them over all embeddings.

Overall, the whole model is trained by minimizing a sum of the standard masked language model loss as in the original BERT design and our new sentence-level language model loss.

\begin{equation}
\label{final_loss}
\mathrm{\mathcal{L}}=\mathcal{L}_{mlm}+\mathcal{L}_{slm}
\end{equation}

Note that the SR module only adds about 7.1\% overhead in computation time to the standard transformer encoder and only during the pre-training stage. It is lightweight compared to the full transformer stack in terms of computation and number of parameters, because it performs re-ordering over a small number of sentential tokens (20 in a 512 token sequence) using shallow decoder stacks (3 layers).

\subsection{Fine-tuning}
The original fine-tuning methods of BERT are slightly modified to encourage the model to use sentence representations learned during pre-training for downstream NLP tasks. Figure~\ref{finetune_arch} shows the overall fine-tuning architecture for extractive QA and classification and regression tasks. 

\vspace{1mm}

\noindent
\textbf{Classification and regression:} The contextualized sentence embeddings of given sentences are concatenated to the [CLS] embedding. Depending on the sentence number of the task, one or two sentence tokens are concatenated before being fed into the output layer. 

\vspace{1mm}

\noindent
\textbf{Extractive QA:} For extractive tasks, we add an answer sentence prediction that finds the sentence token corresponding to the answer, using a pointer network to an existing answer span prediction. To distinguish the question and the context, segment embeddings as in BERT are used while other tasks use the same segment index for all inputs. The final loss is the sum of three independent pointer networks losses: 1) the answer start index, 2) the answer end index, and 3) the answer sentence index.

\begin{table*}[hbt]
\vskip 0.15in
\begin{center}
\begin{small}

\begin{tabular}{lccccccccc}
\hline 
\textbf{Model} & \textbf{CoLA} & \textbf{SST-2} & \textbf{MRPC} & \textbf{STS-B} & \textbf{QQP} & \textbf{MNLI-m/mm} & \textbf{QNLI} & \textbf{RTE} & \textbf{Avg.} \\ 
\hline 
\multicolumn{10}{l}{\textit{Development set results. The best score from hyperparameter searches using parameters in Section~\ref{experiments_setup}.}} \\
BERT\textsubscript{BASE} & 57.3 & 92.9 & 89.0 & 88.6 & 91.4 & 84.8 / 84.9 & 91.8 & 71.5 & 83.6 \\
BERT\textsubscript{LARGE} & 63.1 & 93.2 & 88.0 & 89.5 & 91.7 & 86.6 / 86.6 & 92.3 & 74.0 & 85.0 \\
BERT\textsubscript{LARGE-WWM} & 64.1 & 94.7 & 90.0 & 90.4 & 91.7 & 87.8 / 87.7 & 94.0 & 77.3 & 86.4 \\
\hline 
SLM\textsubscript{BASE}(1M steps) & 62.1 & 93.7 & 90.0 & 90.3 & 91.6 & 86.6 / 86.4 & 93.0 & 81.2 & 86.1 \\
SLM\textsubscript{BASE}(3M steps) & 62.4 & 94.2 & 90.4 & 90.9 & 91.7 & 87.4 / 87.5 & 93.8 & 83.0 & 86.8 \\
\hline 
\multicolumn{10}{l}{\textit{Test set results. Other models' scores are taken from the GLUE leaderboard or their papers}} \\
BERT\textsubscript{BASE} & 52.1 & 93.5 & 88.9 & 87.1 & 89.2 & 84.6 / 83.4 & 90.5 & 66.4 & 81.7 \\
ERNIE 2.0\textsubscript{BASE} & 55.2 & 95.0 & 89.9 & 86.5 & \textbf{89.8} & 86.1 / 85.5 & 92.9 & 74.8 & 84.0 \\
T5\textsubscript{BASE} & 51.1 & \textbf{95.2} & \textbf{90.7} & \textbf{88.6}  & 89.4 & 87.1 / 86.2 & 93.7 & \textbf{80.1} & 84.7 \\
\hline 
BERT\textsubscript{LARGE} & 60.5 & 94.9 & 89.3 & 87.6 & 89.3 & 86.7 / 85.9 & 92.7 & 70.1 & 84.1 \\
XLNet\textsubscript{LARGE} & 63.6 & 95.6 & 89.2 & 91.8 & 91.8 & 89.8 / \hspace{0.23cm}-\hspace{0.23cm} & 93.9 & 83.8 & 87.4 \\
ELECTRA\textsubscript{LARGE} & 68.2 & 94.8 & 89.6 & 91.0  & 90.1 & 90.1 / \hspace{0.23cm}-\hspace{0.23cm} & 95.4 & 83.6 &  87.9  \\
T5\textsubscript{LARGE} & 61.2 & 96.3 & 92.4 & 89.9  & 89.9 & 89.9 / 89.6 & 94.8 & 87.2 & 87.9 \\
\hline 
SLM\textsubscript{BASE}(3M steps) & \textbf{55.3} & 95.1 & 90.0 & 88.3 & 89.6 & \textbf{87.3 / 86.8} & \textbf{93.9} & 78.5 & \textbf{85.0} \\
\hline 
\end{tabular}
\caption{\label{glue-results} GLUE results. BERT dev scores are produced by same hyperparameter searches using the official pre-trained models. Bolded are the highest scores among Base-size models. XLNet and ELECTRA's scores are averages except MNLI-mm that are not reported in the papers.}

\end{small}
\end{center}
\vskip -0.1in
\end{table*}

\section{Experiments}
\label{experiments}

\subsection{Benchmark Datasets}
To measure the excellence of our propose SLM on various downstream tasks, we conduct experiments with three well-known benchmarks: General Language Understanding Evaluation (GLUE), the Stanford Question Answering (SQuAD) v1.1 and v2.0, and Discourse Evaluation (DiscoEval) \cite{mchen-discoeval-19}. We exclude the WNLI task from GLUE following BERT which always achieves 65.1 accuracy of the majority class. RST-DT task is also excluded from DiscoEval which requires a tree structure encoding. Metrics used are Matthew's correlation for CoLA, F1 for MRPC, Spearman's correlation for STS-B, Exact Match (EM) / F1 for SQuAD, and accuracy for rest of tasks. We report development set results for SQuAD to compare with other models, both development and test set scores for GLUE, and test set scores for DiscoEval. 

\subsection{Experimental Setup}
\label{experiments_setup}

For the experiments, we follow BERT\textsubscript{BASE}'s hyperparameters and corpus. Our model is trained with 512 length-256 batch size using Wikipedia dumps and BookCorpus \cite{Zhu_2015_ICCV}. We split inputs into sentences using the NLTK toolkit \cite{Loper02nltk:the}, which are then re-shuffled for every epoch. In terms of data preprocessing, when each input is fed into the model, we set the maximum count $M$ of sentences to process as 20 in our experiments. If exceeded, we randomly merge pairs of adjacent sentences and treat them as a single sentence until the required sentence count is reached. We perform this preprocessing step in order to deal with an uncommonly large number of sentences. For the masked language model objective, we randomly mask up to three continuous word tokens using a geometric distribution, $Geo(p = 0.2)$ following the findings of T5 \cite{Raffel2019ExploringTL} and SpanBERT \cite{joshi2019spanbert}, but sentence tokens are not masked. The models are pre-trained for 1M and 3M steps and optimized by Adam with linear weight decaying using a learning rate 1.5e-4. Any other configuration not mentioned here is the same as the original BERT model. 

Fine-tuning is done by a hyperparameter search of learning rate: \{1e-5, 3e-5, 5e-5, 1e-4\} and epoch between 2 to 15 depends on the tasks. We select the run with the best development set score among five runs for each parameter combination for more stable results. For GLUE test set results, predictions from models with the best development score are submitted. All results of downstream NLP tasks in this paper are results of a single model trained with a single task. For DiscoEval\cite{mchen-discoeval-19} tasks, we freeze the encoder and only fine-tune the output layer following \citet{mchen-discoeval-19}.

Although the recent state-of-the-art models such as ALBERT \cite{lan2019albert} or T5 use extremely large model and training data, we follow BERT\textsubscript{BASE}'s model size because of practical limitations of computation resources. We believe that the BASE settings allow for a robust comparison to the current leading approaches.


\section{Results}
\label{Results}

We compare our SLM model with the original BERT, XLNet \cite{NIPS2019_8812}, ELECTRA \cite{kevin2019electra}, ERNIE 2.0 \cite{sun2019ernie}, CONPONO \cite{Iter2020PretrainingWC}, BART \cite{lewis2019bart} and T5 which are the current state-of-the-art for the benchmark datasets we use. We mainly compare with models using BERT\textsubscript{BASE} hyperparameters.

\subsection{GLUE results}

Table~\ref{glue-results} shows the performance of our method on the GLUE dataset for all tasks except WNLI. Our method significantly improves the downstream NLP tasks in the GLUE dataset. While most tasks are improved from the original BERT, improvement of RTE by 12.1 points is the most significant. We assume this is because our model learns the relationship of sentences thanks to our objective and it is the most effective on the entailment task with relatively small data like RTE. While our average test score is the highest among Base-size models that is 0.3 points higher than T5\textsubscript{BASE}, some scores are even comparable with large models trained by more total tokens and data. Our average score is 0.9 points higher than BERT\textsubscript{LARGE}. Our SST-2 score is higher than BERT and ELECTRA's scores and MRPC score is higher than BERT, XLNet, and ELECTRA.

\subsection{SQuAD results}

\begin{table}
\vspace{0.24cm}
\begin{center}
\begin{small}
\begin{tabular}{lcc}
\hline 
 & \textbf{SQuAD v1.1} & \textbf{SQuAD v2.0} \\
Model & EM / F1 & EM / F1 \\
\hline 
BERT\textsubscript{BASE} & 80.8 / 88.5 & 75.2 / 78.3 \\
XLNet\textsubscript{BASE} & - & 78.5 / 81.3 \\
BART\textsubscript{BASE} &  \hspace{0.15cm} - \hspace{0.15cm} / 90.8 & -\\
ELECTRA\textsubscript{BASE}* & 84.5 / 90.8 & 80.5 / 83.3 \\
T5\textsubscript{BASE} & \textbf{85.4} / 92.1 & -\\
\hline 
BERT\textsubscript{LARGE} & 84.1 / 90.9 & 78.7 / 81.9 \\
BERT\textsubscript{LARGE-WWM} & 86.7 / 92.8 & 82.6 / 85.4 \\
\hline 
SLM\textsubscript{BASE}(1M) & 84.6 / 91.5 & 80.7 / 83.7 \\
SLM\textsubscript{BASE}(3M) & 85.3 / \textbf{92.2} & \textbf{81.9 / 84.9} \\
\hline 
\end{tabular}
\caption{\label{squad-result} Fine-tuning results on the SQuAD v1.1 and v2.0 development sets. The scores of BERT on SQuAD v2.0 are produced using the official pre-trained models and others are taken from their papers. Bolded are the highest scores among Base-size models. *mean scores over 10 runs.}
\end{small}
\end{center}
\vspace{-0.77cm}
\end{table}

As shown in Table~\ref{squad-result}, our method significantly improves the performance on both SQuAD v1.1 and v2.0. It exceeds the original BERT\textsubscript{BASE} model by 3.0 and 5.4 points F1 and 3.8 and 5.5 points EM for v1.1 and v2.0 respectively when it is equally trained for 1M steps. When it is sufficiently trained for 3M steps, additional gains are achieved by 0.7 and 1.2 points EM/F1. Finally, it achieves a tie score with the T5\textsubscript{BASE} model in SQuAD v1.1. This is an impressive result because T5 is an encoder-decoder model which uses twice as many parameters and is trained by 2.7-times more total tokens and astonishingly 37-times more data. 

\begin{table}[h!]
\vspace{0.24cm}
\setlength{\tabcolsep}{4.7pt}
\begin{center}
\begin{small}
\begin{tabular}{lccccc}
\hline 
\textbf{Model} & \textbf{SP} & \textbf{BSO} & \textbf{DC} & \textbf{SSP} & \textbf{PDTB-E/I}  \\
\hline 
BERT\textsubscript{BASE} & 53.1 & 68.5 & 58.9 & 80.3 & 41.9 / 42.4 \\
BERT\textsubscript{LARGE} & 53.8 & 69.3 & 59.6 & 80.4 & 44.3 / 43.6 \\
RoBERTa\textsubscript{BASE} & 38.7 & 58.7 & 58.4 & 79.7 & 39.4 / 40.6 \\
CONPONO\textsubscript{BASE}  & 60.7 & 76.8 & 72.9 & 80.4 & 42.9 / 44.9 \\
\hline 
SLM\textsubscript{BASE}(1M) & 72.4 & 84.1 & 75.4 & \textbf{81.5} & 45.9 / 46.3 \\
SLM\textsubscript{BASE}(3M) & \textbf{73.4} & \textbf{84.5} & \textbf{76.1} & \textbf{81.5}& \textbf{46.4} / \textbf{47.8}\\
\hline 
\end{tabular}
\caption{\label{discoeval-result} Test set results of DiscoEval datasets except RST-DT. Scores of other models are from \citet{mchen-discoeval-19} and \citet{Iter2020PretrainingWC}}
\end{small}
\end{center}
\vspace{-0.77cm}
\end{table}

\subsection{DiscoEval Results}

Table~\ref{discoeval-result} shows the test set results on DiscoEval. Our method achieves improvements with a large margin on all tasks over the previous state-of-the-art model, CONPONO \cite{Iter2020PretrainingWC}. This result reveals that our method learns discourse relations better with our sentence unshuffling objective. Especially, an improvement on SP, finding the original position of a given sentence, is the most significant by 12.7 points higher than CONPONO. This is because sentence positions for a given sentence can be found by finding the next sentence which means the SP task is a reduced problem of our sequence reconstruction task. The improvements on SSP, predicting whether the given sentence belongs to the Abstract section, is relatively little because it has less in common with the SLM objective.


\section{Analysis}
\label{Analysis}

\subsection{Sentence representation in BERT}

\begin{table}[t]
\vskip 0.15in
\begin{center}
\begin{small}
\begin{tabular}{lcc}
\hline 
 & \textbf{SQuAD v1.1} & \textbf{SQuAD v2.0} \\
Model & EM / F1 & EM / F1 \\
\hline 
Original Sentences & 82.7 / 89.8 & 76.5 / 79.4 \\
Shuffled Sentences & \textbf{83.2 / 90.1} & \textbf{77.4 / 80.5} \\
\hline 
\end{tabular}
\caption{\label{bert_compare} The results on SQuAD datasets of models trained with original and shuffled sentences. }
\end{small}
\end{center}
\vskip -0.1in
\end{table} 

To gain motivation for our approach, we begin by running a simple experiment to measure the level of hierarchical understanding of the standard BERT model using the masked language model objective. We train two models: one with sentences in the original ordering and another with the sentences shuffled. The latter model is expected to lose the hierarchical semantics at the discourse level as the shuffling breaks the relations implied by the natural ordering. In practice, the SQuAD results detailed in Table~\ref{bert_compare} show a model trained with the shuffled sentences in fact outperforms the model trained with the sentences in the original ordering. These results indicate that BERT does not take advantage of the global, sentence-level, structure of the text but rather apparently captures a shallower understanding that does not translate into an improvement in reading comprehension performance. 




\begin{table*}[hbt!]

\vskip 0.05in
\begin{center}
\begin{small}
\begin{tabular}{ccccccccccc}
\hline
\begin{tabular}{@{}c@{}}\textbf{Shuf.}\\\textbf{Batch}\end{tabular} &  \begin{tabular}{@{}c@{}}\textbf{Seq.}\\\textbf{Recon.}\end{tabular} & \begin{tabular}{@{}c@{}}\textbf{Sent.}\\\textbf{Repr.}\end{tabular}  &
\begin{tabular}{@{}c@{}}\textbf{SQuAD v1.1}\\\textbf{EM / F1}\end{tabular} & 
\begin{tabular}{@{}c@{}}\textbf{SQuAD v2.0}\\\textbf{EM / F1}\end{tabular} & 
\begin{tabular}{@{}c@{}}\textbf{MNLI}\\\textbf{m / mm}\end{tabular} & 
\textbf{QNLI} & \textbf{SST-2} & \textbf{QQP} & \textbf{Avg.}\\
\hline 
50\% & \checkmark & \checkmark & \textbf{84.6 / 91.5} &  \textbf{80.7 / 83.7} & \textbf{86.6 / 86.4} & \textbf{93.0} & \textbf{93.7}  & \textbf{91.6} & \textbf{88.0}\\
\hline
100\% & \checkmark & \checkmark & 84.2 / 91.2 & 80.3 / 83.4 & 86.1 / 86.2 & 92.9 & 93.5 & 91.4 & 87.7\\
100\% & & \checkmark & 83.7 / 90.7 & 77.3 / 80.4 & 85.4 / 85.6 & 92.3 & 93.0 & 91.3 & 86.6\\
0\% &  &\checkmark & 82.8 / 90.1 & 76.8 / 79.8 & 85.3 / 85.5 & 92.2 & 93.0 & 91.1 & 86.3\\
0\% &  &  & 82.7 / 89.8 & 76.5 / 79.4 & 84.7 / 84.6 & 91.7 & 92.8 & 91.1 & 85.9 \\
\hline
\end{tabular}
\caption{\label{ablation-method} Ablation study on our proposed methods}
\end{small}
\end{center}
\vskip -0.1in
\end{table*}

\subsection{Effect of our proposed method}
In this section, we perform ablation studies to show the impact of our proposed methods. We trained 5 models with different configurations discarding different aspects of the model. We use SQuAD and four tasks from GLUE which are large enough to derive stable comparisons. 

\vspace{1mm}

\noindent
\textbf{Shuffled Batch:} Amount of shuffled batches that need to be balanced between learning by seeing and predicting the original sequence. 

\vspace{1mm}

\noindent
\textbf{Sequence Reconstructor:} Whether the model is trained with a loss from sentence-level objective and the sequence reconstructor model. 

\vspace{1mm}

\noindent
\textbf{Sentence Representation:} Whether the model uses sentence representation tokens and embeddings. 

\vspace{1mm}

While all features increase performance individually, the biggest gains stem from the sequence reconstruction objective with both the shuffled and original orderings. This combination increases SQuAD v1.1 EM / F1 by 1.9 / 1.7 points and v2.0 by 4.2 / 4.3 points respectively. MNLI-m/mm accuracies are similarly improved by 1.9 and 1.8 points. Other tasks are also slightly improved. Notably, our best performing model is achieved by shuffling the sentence ordering only for half of the input batches, to allow the model experiencing also the natural ordering. Removing the ordering from all batches rather than half of them leads to slight reductions of 0.1-0.5 points for most tasks, which might result from the model's lack of exposure to natural input ordering that would be necessary for some tasks. 

One interesting point is that just adding a sentence representation or shuffling the input are each individually increasing the performance even without the sentence-level objective. We conjecture that some of the sentence representations can be learned from the standard word-level prediction loss without the specific objective based on the explicit sentence representation. Shuffling input might help by adding regularization effects and making the model more robust against noisy permutations as well.

\begin{table*}[hbt!]
\vskip 0.15in
\begin{center}
\begin{small}
\begin{tabular}{p{12mm}p{137mm}} 
\hline 
\textbf{Discourse} & \textbf{Sources and Nearest Neighbors}    \\
\hline 

Contrast & 
\textbf{Q} : \textnormal{The waterwheel hammered on. \textbf{\textcolor{red}{Otherwise} there was \textcolor{blue}{silence}.}  } \\
 \multicolumn{1}{r}{$\times$} &\textnormal{1) A portion of the burning log fell on to the hearth. \textbf{Then there was \textcolor{blue}{silence}}}  \\
&\textnormal{2) The bees still worked on, and the butterflies did not rest from roving, their smallness seeming to shield them from the stagnating effect that this turning moment of day had on larger creatures. \textbf{\textcolor{red}{Otherwise} \textcolor{blue}{all was still}.}}   \\ 
&\textnormal{3) A class in spelling, big boys and little girls, toed a crack in front of the waster's desk. \textbf{\textcolor{red}{The rest of} the school droned away on appointed tasks in the drowsy interlude.}}\\
\multicolumn{1}{r}{$\times$}&\textnormal{4) Fra Girolamo, give her-the Crucifix, said the voice of Fra Girolamo. \textbf{\textcolor{blue}{No other sound came from} the dying lips. }} \\
\hline 

Strengthen & 
\textbf{Q} : \textnormal{It doesn't hurt at all. \textbf{\textcolor{red}{In fact} it's exhilarating.}}  \\
&\textnormal{1) We have no great political guns aboard. \textbf{\textcolor{red}{On the contrary}, the majority of the passengers are Americans.}} \\
&\textnormal{2) I don't believe in the faults. \textbf{They're just a joyous softening of the outline - more beautiful than perfection.}} \\
&\textnormal{3) I never regarded Ealer's readings as educational. \textbf{\textcolor{red}{Indeed}, they were a detriment to me.}} \\
&\textnormal{7) We have no sympathy at all with the moral indignation of our time against M. Zola. \textbf{It is simply the indignation of Tartuffe on being exposed.}} \\
\hline 

Elaboration & 
\textbf{Q} : \textnormal{Then I became a woman. \textbf{A strong one at that.}} \\
&
\textnormal{1) Christianity was the last great religious synthesis. \textbf{It is the one nearest to us.}} \\
&\textnormal{2) I wanted you to care for me so that I could influence you. \textbf{It wasn't easy.} } \\
&\textnormal{3) It got on my nerves - the women I saw. \textbf{Worse than any man.}} \\
&\textnormal{9) He's past his seventy now, - ever so much; but he's just as modest as a young girl. \textbf{A deal more modest than some of them.}
} \\
\hline 
Return & 
\textbf{Q} : \textnormal{He had a point. \textbf{\textcolor{red}{Still} for good measure, I pouted.}}  \\
& \textnormal{1) Logically - if not legally - there is apparently an inference of the interchange of matrimonial consent here. \textbf{I stick to my own opinion, \textcolor{red}{nevertheless}.}}\\
&\textnormal{2) To him I was a squeezed lemon. \textbf{\textcolor{red}{Nevertheless} I took his hint.}}  \\
&\textnormal{3) It was almost dark. \textbf{\textcolor{red}{Yet} I must walk away.} } \\
&\textnormal{9) He spoke not a word. \textbf{I pitied him from the bottom of my heart.}
} \\
\hline 

Temporal & 
\textbf{Q} : \textnormal{It limped closer at a slow pace. \textbf{\textcolor{red}{Soon} it \textcolor{blue}{stopped} in front of us.}}  \\
& \textnormal{1) The current carried them on and on, but not so swiftly as it was carrying the tree. \textbf{\textcolor{red}{Soon} they were approaching the bend.} } \\
& \textnormal{2) Then he got down from his post and loafed along the sidewalk, still observing and occasionally commenting. \textbf{\textcolor{red}{Presently} he dropped into my wake and followed along behind.} } \\
& \textnormal{3) Slowly, patiently, watchfully, the hunter followed. \textbf{\textcolor{red}{After a while} he \textcolor{blue}{stopped} with a satisfied grin.} } \\
& \textnormal{12) The mass fell into columns by threes and fours to accommodate itself to the narrow road, and strode briskly along southward in the wake of the leaders. 	 \textbf{\textcolor{red}{In a few minutes} the Hogan cabin was \textcolor{blue}{reached}.} 
} \\
\hline 

List & \textbf{Q} : \textnormal{I saw flowers on the ground. \textbf{I \textcolor{blue}{heard} birds in the trees.}} \\
\multicolumn{1}{r}{$\times$}& \textnormal{1) He saw a garden. \textbf{We saw a wilderness.}}  \\
& \textnormal{2) I heard the pulse of the besieging sea throb far away all night. \textbf{I \textcolor{blue}{heard} the wind fly crying, and convulse tumultuous palms.} } \\
\multicolumn{1}{r}{$\times$}& \textnormal{3) I took several long walks while collecting objects of natural history. 	 \textbf{The country is pleasant for exercise.} } \\
& \textnormal{5) My maid is a treasure. \textbf{My dressmaker is charming.} } \\
\hline 
\end{tabular}
\caption{\label{nearest_with_discourse} Nearest neighbors of contextualized sentence representations. }
\end{small}
\end{center}
\vskip -0.1in
\end{table*}

\subsection{What information is learned by SLM?}

For a more intuitive understanding of what is learned from the unshuffling objective, we search the closest sentences to query sentences using a cosine distance of sentence representations from our pre-trained model. Sentences with typical discourse labels from samples in \citeauthor{jernite2017discoursebased} \citeyearpar{jernite2017discoursebased} are used as queries to see whether our model captures a variety of different discourses and 1M sentence pairs from Gutenberg BookCorpus \cite{lahiri} are used as targets to search. 
Table~\ref{nearest_with_discourse} shows the Top 3 closest sentences for each query sentence with discourse labels. We additionally pick one of the highly ranked results to show diversities of similarities between queries and retrievals. We mark query and retrieved sentences in bold and show previous sentences together. Semantically similar phrases are marked in blue while potential clues to discourse relations are in red. Results that do not match the query's discourse label are marked by $\times$.

As we can see from the table, most of the retrieved sentences share a mixture of syntactic, semantic and discourse-level aspects with the given queries, especially in regard to their relationships with their surrounding context. Even without any fine-tuning and filtering of the retrieved results, most of the highly ranked ones share a similar discourse relation with their prior sentences to that of the query sentences. Some discourse relations are commonly indicated by conjunctive words, and then the retrieved sentences have conjunctions of the relevant discourse sense, e.g. \textit{``in fact''}, and \textit{``soon''} for \textit{strengthen}, and words indicating similar temporal relations for \textit{temporal}. In other cases, when the results differ in the conjunctive words, they still hold high-level structural similarities to the query sentences. This is especially true for the \textit{elaboration} and \textit{list} relations. Finally, we notice that semantically similar phrases are also captured by the retrieved results. In the \textit{contrast} example, results contain phrases like \textit{``all was still''}, and \textit{``no other sound came from''} which correspond to the word \textit{``silence''} in the query. It is noticeable that overall our model seems to focus more on the relational aspect than the semantic and syntactic aspects. This serves as a qualitative evidence for the effectiveness of the new unshuffling objective: while all aspects are needed for sentence understanding, we conjecture that the relational component is the one most vital for the un-shuffling task as it naturally demands an understanding of the relationships between the sentences.

\section{Related Work}
\label{related_work}

\textbf{Contextualized Representation learning for NLP:} Self-supervised representation learning became popular after contextualization methods were introduced. ELMo \cite{Peters:2018} dynamically contextualizes representations of adjacent words using bi-directional recurrent encoders. BERT \cite{devlin2018bert} adopted a deep transformer encoder and has proposed the masked language modeling objective incorporating a bi-directional context. After the success of BERT, researchers started exploring various new pre-training objectives such as span boundary representations \cite{joshi2019spanbert}, reordering of local permutations \cite{wang2019structbert}, detecting incorrectly replaced tokens \cite{kevin2019electra}, combining multiple tasks \cite{sun2019ernie}, a decoder-based masked word prediction \cite{song2019mass}, and so on. 

Another line of works tried to scale up the model with more parameters, training data, and computation resources. RoBERTa \cite{liu2019roberta} trains BERT\textsubscript{LARGE} model with 10x more data for 16x more iterations. ALBERT \cite{lan2019albert} uses the xxlarge model whose parameters are reduced by weight sharing and factorizations. T5 proposes a multi-task encoder-decoder architecture that uses up-to about 33x weight parameters, 37x data size compared to the original BERT\textsubscript{LARGE} model. 

\vspace{1mm}

\noindent
\textbf{Sentence Representation Models:} 
Several works tried to learn sentence representations using adjacent sentences, especially in recurrent networks. SkipThoughts \cite{kiros2015skip} and FastSent \cite{hill2016learning} proposed an encoder-decoder architecture that encodes a sentence and generates the next and previous sentences. These works focus on independent representations of single independent sentences and do not consider the dynamic contextualization using neighboring sentences.

We believe that, similar to words, sentence representations should also be properly contextualized in order to embed richer meaning including relationships to other sentences. HLSTM \cite{Chang2019LanguageMP} considered contextualization of sentence representations by incorporating previous step's sentence representation for word prediction. HIBERT \cite{zhang-etal-2019-hibert2} proposed a hierarchical transformer encoder trained by recovering masked sentences, focusing in particular on summarization. However, both of these models are trained by performing prediction at the word level, wheres our sentence unshuffling approach demands fine understanding of the relations among the sentences at the more global discourse level. 

\vspace{1mm}

\noindent
\textbf{Learning from Sentence Ordering:} There are prior works about learning contextualized sentence representations by recovering the original sequence of sentences. \citet{gong2016end} and \citet{logeswaran2018sentence} proposed RNN-based pointer networks for reconstructing the sequence order. However, they utilized hierarchical models that encode each sentence separately without any access to other sentences. This fact not only restricts long-term contextualization of words and sentences but also limits downstream tasks due to the dedicated architectures. On the other hand, our method encodes multiple sentence representations at the same time with both inner and inter-sentence contextualization. 

The effect of predicting textual segment order in pretrained language models has been widely investigated as well. BERT proposed the Next Sentence Prediction task (NSP) which predicts whether two given text segments are from the same documents or not. The results of SpanBERT \cite{joshi2019spanbert} question the value of NSP, suggesting this might be due to noise from merging 2 unrelated texts from different documents. Consequently, ALBERT \cite{lan2019albert} and StructBERT \cite{wang2019structbert} added sentence ordering objectives by predicting the order of text segments. BART \cite{lewis2019bart} proposed the Sentence Permutation task which is similar with ours. It predicts the original sequence of sentences using an auto-regressive decoder, which reconstructs the whole sentences by word prediction. However, their approach does not provide any representation of each sentence. Moreover, while SLM shows strong task improvements, that is not the case for these models.

\section{Conclusions}
\label{conclusions}

In this paper, we proposed the Sentence-level Language Modeling objective for contextualized sentence representation learning. Our approach extends a word-level language modeling strategy to the sentence-level by reconstructing the original order of shuffled sentences. In addition, we designed a special Sequence Reconstructor (SR) module to learn to perform the sentence re-ordering. It reconstructs the original order by pointing to the next sentence among encoded sentence representations using a pointer network and a transformer decoder. We evaluated the effect of the proposed idea on three benchmarks, GLUE, SQuAD, and DiscoEval, and showed consistent improvements over the previous approaches. We matched performance with the state-of-the-art model with the same model size using much fewer parameters, computation, and data. Through a qualitative analysis, we showed that our model can embed not only semantic but also relational features of sentences. We are excited about future work that could extend our motivation and further aim at incorporating stronger hierarchy into the language model architectures and the pre-training tasks.

\bibliographystyle{acl_natbib}
\bibliography{main}

\begin{thebibliography}{33}
\expandafter\ifx\csname natexlab\endcsname\relax\def\natexlab#1{#1}\fi

\bibitem[{Abadi et~al.(2015)Abadi, Agarwal, Barham, Brevdo, Chen, Citro,
  Corrado, Davis, Dean, Devin, Ghemawat, Goodfellow, Harp, Irving, Isard, Jia,
  Jozefowicz, Kaiser, Kudlur, Levenberg, Man\'{e}, Monga, Moore, Murray, Olah,
  Schuster, Shlens, Steiner, Sutskever, Talwar, Tucker, Vanhoucke, Vasudevan,
  Vi\'{e}gas, Vinyals, Warden, Wattenberg, Wicke, Yu, and
  Zheng}]{tensorflow2015-whitepaper}
Mart\'{\i}n Abadi, Ashish Agarwal, Paul Barham, Eugene Brevdo, Zhifeng Chen,
  Craig Citro, Greg~S. Corrado, Andy Davis, Jeffrey Dean, Matthieu Devin,
  Sanjay Ghemawat, Ian Goodfellow, Andrew Harp, Geoffrey Irving, Michael Isard,
  Yangqing Jia, Rafal Jozefowicz, Lukasz Kaiser, Manjunath Kudlur, Josh
  Levenberg, Dandelion Man\'{e}, Rajat Monga, Sherry Moore, Derek Murray, Chris
  Olah, Mike Schuster, Jonathon Shlens, Benoit Steiner, Ilya Sutskever, Kunal
  Talwar, Paul Tucker, Vincent Vanhoucke, Vijay Vasudevan, Fernanda Vi\'{e}gas,
  Oriol Vinyals, Pete Warden, Martin Wattenberg, Martin Wicke, Yuan Yu, and
  Xiaoqiang Zheng. 2015.
\newblock \href {https://www.tensorflow.org/} {{TensorFlow}: Large-scale
  machine learning on heterogeneous systems}.
\newblock Software available from tensorflow.org.

\bibitem[{Chang et~al.(2019)Chang, Toutanova, Lee, and
  Devlin}]{Chang2019LanguageMP}
Ming-Wei Chang, Kristina Toutanova, Kenton Lee, and Jacob Devlin. 2019.
\newblock Language model pre-training for hierarchical document
  representations.
\newblock \emph{arXiv preprint arXiv:1901.09128}.

\bibitem[{Chen et~al.(2019)Chen, Chu, and Gimpel}]{mchen-discoeval-19}
Mingda Chen, Zewei Chu, and Kevin Gimpel. 2019.
\newblock Evaluation benchmarks and learning criteria for discourse-aware
  sentence representations.
\newblock In \emph{Proceedings of the 2019 Conference on Empirical Methods in
  Natural Language Processing and the 9th International Joint Conference on
  Natural Language Processing (EMNLP-IJCNLP)}, pages 649--662.

\bibitem[{Chen et~al.(2016)Chen, Qiu, and Huang}]{chen2016neural}
Xinchi Chen, Xipeng Qiu, and Xuanjing Huang. 2016.
\newblock Neural sentence ordering.
\newblock \emph{arXiv preprint arXiv:1607.06952}.

\bibitem[{Clark et~al.(2019)Clark, Luong, Le, and Manning}]{kevin2019electra}
Kevin Clark, Minh-Thang Luong, Quoc~V Le, and Christopher~D Manning. 2019.
\newblock Electra: Pre-training text encoders as discriminators rather than
  generators.
\newblock In \emph{International Conference on Learning Representations}.

\bibitem[{Devlin et~al.(2019)Devlin, Chang, Lee, and
  Toutanova}]{devlin2018bert}
Jacob Devlin, Ming-Wei Chang, Kenton Lee, and Kristina Toutanova. 2019.
\newblock Bert: Pre-training of deep bidirectional transformers for language
  understanding.
\newblock In \emph{Proceedings of the 2019 Conference of the North American
  Chapter of the Association for Computational Linguistics: Human Language
  Technologies, Volume 1 (Long and Short Papers)}, pages 4171--4186.

\bibitem[{Gan et~al.(2017)Gan, Pu, Henao, Li, He, and Carin}]{gan2017learning}
Zhe Gan, Yunchen Pu, Ricardo Henao, Chunyuan Li, Xiaodong He, and Lawrence
  Carin. 2017.
\newblock Learning generic sentence representations using convolutional neural
  networks.
\newblock In \emph{Proceedings of the 2017 Conference on Empirical Methods in
  Natural Language Processing}, pages 2390--2400.

\bibitem[{Gong et~al.(2016)Gong, Chen, Qiu, and Huang}]{gong2016end}
Jingjing Gong, Xinchi Chen, Xipeng Qiu, and Xuanjing Huang. 2016.
\newblock End-to-end neural sentence ordering using pointer network.
\newblock \emph{arXiv preprint arXiv:1611.04953}.

\bibitem[{Hill et~al.(2016)Hill, Cho, and Korhonen}]{hill2016learning}
Felix Hill, Kyunghyun Cho, and Anna Korhonen. 2016.
\newblock Learning distributed representations of sentences from unlabelled
  data.
\newblock In \emph{Proceedings of the 2016 Conference of the North American
  Chapter of the Association for Computational Linguistics: Human Language
  Technologies}, pages 1367--1377.

\bibitem[{Iter et~al.(2020)Iter, Guu, Lansing, and
  Jurafsky}]{Iter2020PretrainingWC}
Dan Iter, Kelvin Guu, Larry Lansing, and Dan Jurafsky. 2020.
\newblock Pretraining with contrastive sentence objectives improves discourse
  performance of language models.
\newblock In \emph{Proceedings of the 58th Annual Meeting of the Association
  for Computational Linguistics}, pages 4859--4870.

\bibitem[{Jernite et~al.(2017)Jernite, Bowman, and
  Sontag}]{jernite2017discoursebased}
Yacine Jernite, Samuel~R Bowman, and David Sontag. 2017.
\newblock Discourse-based objectives for fast unsupervised sentence
  representation learning.
\newblock \emph{arXiv preprint arXiv:1705.00557}.

\bibitem[{Joshi et~al.(2020)Joshi, Chen, Liu, Weld, Zettlemoyer, and
  Levy}]{joshi2019spanbert}
Mandar Joshi, Danqi Chen, Yinhan Liu, Daniel~S Weld, Luke Zettlemoyer, and Omer
  Levy. 2020.
\newblock Spanbert: Improving pre-training by representing and predicting
  spans.
\newblock \emph{Transactions of the Association for Computational Linguistics},
  8:64--77.

\bibitem[{Kiros et~al.(2015)Kiros, Zhu, Salakhutdinov, Zemel, Urtasun,
  Torralba, and Fidler}]{kiros2015skip}
Ryan Kiros, Yukun Zhu, Russ~R Salakhutdinov, Richard Zemel, Raquel Urtasun,
  Antonio Torralba, and Sanja Fidler. 2015.
\newblock Skip-thought vectors.
\newblock In \emph{Advances in neural information processing systems}, pages
  3294--3302.

\bibitem[{Lahiri(2014)}]{lahiri}
Shibamouli Lahiri. 2014.
\newblock Complexity of word collocation networks: A preliminary structural
  analysis.
\newblock In \emph{Proceedings of the Student Research Workshop at the 14th
  Conference of the European Chapter of the Association for Computational
  Linguistics}, pages 96--105.

\bibitem[{Lan et~al.(2019)Lan, Chen, Goodman, Gimpel, Sharma, and
  Soricut}]{lan2019albert}
Zhenzhong Lan, Mingda Chen, Sebastian Goodman, Kevin Gimpel, Piyush Sharma, and
  Radu Soricut. 2019.
\newblock Albert: A lite bert for self-supervised learning of language
  representations.
\newblock In \emph{International Conference on Learning Representations}.

\bibitem[{Lewis et~al.(2020)Lewis, Liu, Goyal, Ghazvininejad, Mohamed, Levy,
  Stoyanov, and Zettlemoyer}]{lewis2019bart}
Mike Lewis, Yinhan Liu, Naman Goyal, Marjan Ghazvininejad, Abdelrahman Mohamed,
  Omer Levy, Ves Stoyanov, and Luke Zettlemoyer. 2020.
\newblock Bart: Denoising sequence-to-sequence pre-training for natural
  language generation, translation, and comprehension.
\newblock In \emph{Proceedings of the 58th Annual Meeting of the Association
  for Computational Linguistics}, pages 7871--7880.

\bibitem[{Liu et~al.(2019)Liu, Ott, Goyal, Du, Joshi, Chen, Levy, Lewis,
  Zettlemoyer, and Stoyanov}]{liu2019roberta}
Yinhan Liu, Myle Ott, Naman Goyal, Jingfei Du, Mandar Joshi, Danqi Chen, Omer
  Levy, Mike Lewis, Luke Zettlemoyer, and Veselin Stoyanov. 2019.
\newblock Roberta: A robustly optimized bert pretraining approach.
\newblock \emph{arXiv preprint arXiv:1907.11692}.

\bibitem[{Logeswaran et~al.(2018)Logeswaran, Lee, and
  Radev}]{logeswaran2018sentence}
Lajanugen Logeswaran, Honglak Lee, and Dragomir~R Radev. 2018.
\newblock Sentence ordering and coherence modeling using recurrent neural
  networks.
\newblock In \emph{AAAI}.

\bibitem[{Loper and Bird(2002)}]{Loper02nltk:the}
Edward Loper and Steven Bird. 2002.
\newblock Nltk: The natural language toolkit.
\newblock In \emph{Proceedings of the ACL-02 Workshop on Effective Tools and
  Methodologies for Teaching Natural Language Processing and Computational
  Linguistics}, pages 63--70.

\bibitem[{Peters et~al.(2018)Peters, Neumann, Iyyer, Gardner, Clark, Lee, and
  Zettlemoyer}]{Peters:2018}
Matthew~E Peters, Mark Neumann, Mohit Iyyer, Matt Gardner, Christopher Clark,
  Kenton Lee, and Luke Zettlemoyer. 2018.
\newblock Deep contextualized word representations.
\newblock In \emph{Proceedings of NAACL-HLT}, pages 2227--2237.

\bibitem[{Raffel et~al.(2020)Raffel, Shazeer, Roberts, Lee, Narang, Matena,
  Zhou, Li, and Liu}]{Raffel2019ExploringTL}
Colin Raffel, Noam Shazeer, Adam Roberts, Katherine Lee, Sharan Narang, Michael
  Matena, Yanqi Zhou, Wei Li, and Peter~J. Liu. 2020.
\newblock Exploring the limits of transfer learning with a unified text-to-text
  transformer.
\newblock \emph{Journal of Machine Learning Research}, 21(140):1--67.

\bibitem[{Rajpurkar et~al.(2018)Rajpurkar, Jia, and Liang}]{rajpurkar2018know}
Pranav Rajpurkar, Robin Jia, and Percy Liang. 2018.
\newblock Know what you don’t know: Unanswerable questions for squad.
\newblock In \emph{Proceedings of the 56th Annual Meeting of the Association
  for Computational Linguistics (Volume 2: Short Papers)}, pages 784--789.

\bibitem[{Reddy et~al.(2019)Reddy, Chen, and
  Manning}]{DBLP:journals/corr/abs-1808-07042}
Siva Reddy, Danqi Chen, and Christopher~D Manning. 2019.
\newblock Coqa: A conversational question answering challenge.
\newblock \emph{Transactions of the Association for Computational Linguistics},
  7:249--266.

\bibitem[{Sergeev and Balso(2018)}]{sergeev2018horovod}
Alexander Sergeev and Mike~Del Balso. 2018.
\newblock Horovod: fast and easy distributed deep learning in {TensorFlow}.
\newblock \emph{arXiv preprint arXiv:1802.05799}.

\bibitem[{Song et~al.(2019)Song, Tan, Qin, Lu, and Liu}]{song2019mass}
Kaitao Song, Xu~Tan, Tao Qin, Jianfeng Lu, and Tie-Yan Liu. 2019.
\newblock Mass: Masked sequence to sequence pre-training for language
  generation.
\newblock In \emph{International Conference on Machine Learning}, pages
  5926--5936.

\bibitem[{Sun et~al.(2020)Sun, Wang, Li, Feng, Tian, Wu, and
  Wang}]{sun2019ernie}
Y.~Sun, Shuohuan Wang, Yukun Li, Shikun Feng, Hao Tian, H.~Wu, and Haifeng
  Wang. 2020.
\newblock Ernie 2.0: A continual pre-training framework for language
  understanding.
\newblock In \emph{AAAI}.

\bibitem[{Vaswani et~al.(2017)Vaswani, Shazeer, Parmar, Uszkoreit, Jones,
  Gomez, Kaiser, and Polosukhin}]{transformer}
Ashish Vaswani, Noam Shazeer, Niki Parmar, Jakob Uszkoreit, Llion Jones,
  Aidan~N Gomez, {\L}ukasz Kaiser, and Illia Polosukhin. 2017.
\newblock Attention is all you need.
\newblock In \emph{Advances in neural information processing systems}, pages
  5998--6008.

\bibitem[{Wang et~al.(2018)Wang, Singh, Michael, Hill, Levy, and
  Bowman}]{DBLP:journals/corr/abs-1804-07461}
Alex Wang, Amanpreet Singh, Julian Michael, Felix Hill, Omer Levy, and Samuel
  Bowman. 2018.
\newblock Glue: A multi-task benchmark and analysis platform for natural
  language understanding.
\newblock In \emph{Proceedings of the 2018 EMNLP Workshop BlackboxNLP:
  Analyzing and Interpreting Neural Networks for NLP}, pages 353--355.

\bibitem[{Wang et~al.(2019)Wang, Bi, Yan, Wu, Xia, Bao, Peng, and
  Si}]{wang2019structbert}
Wei Wang, Bin Bi, Ming Yan, Chen Wu, Jiangnan Xia, Zuyi Bao, Liwei Peng, and
  Luo Si. 2019.
\newblock Structbert: Incorporating language structures into pre-training for
  deep language understanding.
\newblock In \emph{International Conference on Learning Representations}.

\bibitem[{Wu et~al.(2016)Wu, Schuster, Chen, Le, Norouzi, Macherey, Krikun,
  Cao, Gao, Macherey, Klingner, Shah, Johnson, Liu, Łukasz Kaiser, Gouws,
  Kato, Kudo, Kazawa, Stevens, Kurian, Patil, Wang, Young, Smith, Riesa,
  Rudnick, Vinyals, Corrado, Hughes, and Dean}]{wu2016googles}
Yonghui Wu, Mike Schuster, Zhifeng Chen, Quoc~V. Le, Mohammad Norouzi, Wolfgang
  Macherey, Maxim Krikun, Yuan Cao, Qin Gao, Klaus Macherey, Jeff Klingner,
  Apurva Shah, Melvin Johnson, Xiaobing Liu, Łukasz Kaiser, Stephan Gouws,
  Yoshikiyo Kato, Taku Kudo, Hideto Kazawa, Keith Stevens, George Kurian,
  Nishant Patil, Wei Wang, Cliff Young, Jason Smith, Jason Riesa, Alex Rudnick,
  Oriol Vinyals, Greg Corrado, Macduff Hughes, and Jeffrey Dean. 2016.
\newblock \href {http://arxiv.org/abs/1609.08144} {Google's neural machine
  translation system: Bridging the gap between human and machine translation}.

\bibitem[{Yang et~al.(2019)Yang, Dai, Yang, Carbonell, Salakhutdinov, and
  Le}]{NIPS2019_8812}
Zhilin Yang, Zihang Dai, Yiming Yang, Jaime Carbonell, Russ~R Salakhutdinov,
  and Quoc~V Le. 2019.
\newblock Xlnet: Generalized autoregressive pretraining for language
  understanding.
\newblock In \emph{Advances in neural information processing systems}, pages
  5753--5763.

\bibitem[{Zhang et~al.(2019)Zhang, Wei, and Zhou}]{zhang-etal-2019-hibert2}
Xingxing Zhang, Furu Wei, and Ming Zhou. 2019.
\newblock Hibert: Document level pre-training of hierarchical bidirectional
  transformers for document summarization.
\newblock In \emph{Proceedings of the 57th Annual Meeting of the Association
  for Computational Linguistics}, pages 5059--5069.

\bibitem[{Zhu et~al.(2015)Zhu, Kiros, Zemel, Salakhutdinov, Urtasun, Torralba,
  and Fidler}]{Zhu_2015_ICCV}
Yukun Zhu, Ryan Kiros, Rich Zemel, Ruslan Salakhutdinov, Raquel Urtasun,
  Antonio Torralba, and Sanja Fidler. 2015.
\newblock Aligning books and movies: Towards story-like visual explanations by
  watching movies and reading books.
\newblock In \emph{Proceedings of the IEEE international conference on computer
  vision}, pages 19--27.

\end{thebibliography}

\clearpage
\appendix


\def\thesection{\Alph{section}}
\section{Training Details}
\label{supp}

We provide details of hyperparameters and corpus we used for pre-training and fine-tunings. Table~\ref{pretrain-hyperparam} shows all the detailed hyperparamters. We mainly used BERT-Base size (12 layers, 12 attention headers, 768 hidden size, and 110M parameters). Pre-training a BERT-Large sized model for 3M iterations is expected to take more than a month in our experiment environments, so we stick to BERT-Base size. All other hyperparameters except the learning rate during pre-training are the same as the original BERT. We found that LR 1.5e-4 results in slightly better compare to LR 1e-4 used in the original BERT.

\begin{table}[h]

\vskip 0.15in
\begin{center}
\begin{small}
\begin{tabular}{lcc}
\toprule
Parameter & Pre-training & Fine-tuning \\
\midrule
Encoder layers & 12 & 12 \\
Decoder layers & 3 & - \\
Attention heads & 12 & 12 \\
Hidden size & 768 & 768 \\
Sequence length & 512 & 128 / 512 \\
Sentence tokens & 20 & 20 \\
Vocab Size & 30522 & 30522 \\
\midrule
Batch size & 256 & 32 \\
Step / Epoch & 1M/3M & $2 \sim{} 15$ \\
Warm-up & 10K & 10\% \\
\midrule
Learning Rate & 1.5e-4  & 1/3/5e-5/1e-4 \\
Adam epsilon & 1e-6 & 1e-6 \\
Dropout & 0.1 & 0.1 \\
Attention Dropout & 0.1 & 0.1 \\
Precision & Float16 & Float16 \\
\bottomrule
\end{tabular}
\caption{\label{pretrain-hyperparam} Training Hyperparameters.}
\end{small}
\end{center}
\vskip -0.1in
\end{table}

We use English Wikipedia and Bookcorpus\cite{Zhu_2015_ICCV} for pre-training. Since Bookcorpus is no longer available online, we collected our own version of BookCorpus using publicly available crawling code (\url{https://github.com/soskek/bookcorpus}). Texts are extracted from Wikipedia dumps using WikiExtractor (\url{https://github.com/attardi/wikiextractor}). Simple heuristic preprocesses are applied to clean the corpus which removed sentences that do not contain enough words. The final size of processed corpus are 9.2GB and 5.7 GB for Wikipedia and BookCorpus respectively. The total size is slightly smaller compare to the total corpus used to train the original BERT because some links are unavailable when we collected the BookCorpus. We used 8 NVIDIA v100 GPUs for pretraining and it took about 5 days to train for 1M iteration. We tokenize the corpus with WordPiece \cite{wu2016googles} tokenizer using uncased vocabulary of Google's official BERT release.

For all fine-tuning tasks, we use batch size 32 with 512 sequence length for SQuAD and 128 for GLUE tasks. Hyperparameter searches are mainly done for epoch \{2, 3, 4, 5\} and learning rate \{ 1e-5, 3e-5, 5e-5 \}. We run 5 runs for each parameter combinations to get a stable dev set results. GLUE test set scores are achieved by submitting test set predictions of the best dev score models to the glue evaluation server. We report the best hyperparameters for each task in Table~\ref{finetune-best-hyperparam}.

\begin{table}[h]

\vskip 0.15in
\begin{center}
\begin{small}
\begin{tabular}{lcc}
\toprule
Task & Epoch & Learning Rate \\
\midrule
SQuAD v1.1 & 2 & 5e-5 \\
SQuAD v2.0 & 2 & 3e-5 \\
\midrule
CoLA & 4 & 1e-5 \\
SST-2 & 3 & 3e-5 \\
MRPC & 3 & 3e-5 \\
STS-B & 4 & 1e-5 \\
QQP & 4 & 3e-5 \\
MNLI & 2 & 3e-5 \\
QNLI & 2 & 3e-5 \\
RTE & 5 & 1e-5 \\
\midrule
SP & 13 & 1e-4 \\
BSO & 13 & 5e-5 \\
DC & 15 & 5e-5\\
SPP & 10 & 1e-4 \\
PDTB-E & 15 & 1e-4\\
PDTB-I & 15 & 1e-4\\
\bottomrule
\end{tabular}
\caption{\label{finetune-best-hyperparam} Best hyperparameters for NLP tasks.}
\end{small}
\end{center}
\vskip -0.1in
\end{table}

\section{Implementation Details}
We use Tensorflow\cite{tensorflow2015-whitepaper} for all of our experiments. Our implementations are based on the NVIDIA's tensorflow implementation of BERT which supports multi GPU using Horovod \cite{sergeev2018horovod} and half precision training  (\url{https://github.com/NVIDIA/DeepLearningExamples/tree/master/TensorFlow/LanguageModeling/BERT}). We adapt an implementation of transformer decoder from official tensorflow's transformer implementation  (\url{https://github.com/tensorflow/models/tree/master/official/transformer})

\end{document}